\documentclass[10pt, journal]{IEEEtran}

\makeatletter
\def\ps@headings{%
\def\@oddhead{\mbox{}\scriptsize\rightmark \hfil \thepage}%
\def\@evenhead{\scriptsize\thepage \hfil \leftmark\mbox{}}%
\def\@oddfoot{}%
\def\@evenfoot{}}
\makeatother
\usepackage[T1]{fontenc}
\usepackage[utf8]{inputenc}
\usepackage[euler]{textgreek}

\usepackage{amsmath}
\usepackage{amssymb}
\usepackage{amsfonts}
\usepackage{amsthm}
\usepackage{bm}

\usepackage[pdftex]{graphicx}
\DeclareGraphicsExtensions{.pdf,.jpeg,.png,.eps}
\usepackage{epsfig}
\usepackage{epstopdf}

\usepackage[table,xcdraw]{xcolor}
\definecolor{darkgreen}{rgb}{0.01, 0.75, 0.24}
\definecolor{usethiscolorhere}{rgb}{0.86666,0.78431,0.78431}
\definecolor{lime}{HTML}{A6CE39}
\definecolor{headingcolor}{rgb}{0.3,0.6,0.8}
\definecolor{controlcolor}{rgb}{0.8,0.4,0.4}
\definecolor{subcontrolcolor}{rgb}{0.6,0.2,0.2}

\usepackage{multirow}
\usepackage{multicol}
\usepackage{array}
\usepackage{tabularx}
\usepackage{booktabs}
\usepackage{colortbl}
\usepackage{hhline}
\usepackage{arydshln}
\usepackage{threeparttable}
\usepackage{adjustbox}
\usepackage{rotating}
\usepackage{dblfloatfix}

\newcolumntype{L}{>{\arraybackslash}m{5.7cm}}
\newcolumntype{I}{>{\arraybackslash}m{5cm}}
\newcolumntype{J}{>{\arraybackslash}m{8cm}}
\newcolumntype{K}{>{\arraybackslash}m{7cm}}
\newcolumntype{M}{>{\arraybackslash}m{3cm}}

\usepackage{float}
\usepackage{placeins}
\usepackage{caption}
\usepackage{subcaption}
\usepackage{algorithm}
\usepackage{algorithmic}

\usepackage{enumitem}

\usepackage{setspace}
\usepackage{soul}
\usepackage{csquotes}
\usepackage{dirtytalk}
\usepackage{verbatim}
\usepackage{comment}
\usepackage[hyphens]{url}

\usepackage[colorlinks,breaklinks]{hyperref}

\usepackage{tikz}
\usetikzlibrary{mindmap}
\usetikzlibrary{calc,positioning}
\usetikzlibrary{decorations.pathmorphing}

\usepackage{lipsum}
\usepackage{balance}

\usepackage{pifont}
\newcommand{\cmark}{\ding{51}}
\newcommand{\xmark}{\ding{55}}

\usepackage[colorinlistoftodos]{todonotes}

\begin{document}
\title{Smart Railway Obstruction Detection System using IoT and Computer Vision}

\author{Pravin Kumar, Mritunjay Shall Peelam, Ramakant Kumar, Sanjay Kumar, Vinay Chamola \textit{Senior Member, IEEE},

\thanks{Pravin Kumar and Mritunjay Shall Peelam are with School of Computer Science, University of Petroleum and Energy Studies (UPES), Dehradun, 248001, Uttarakhand, India. (e-mail: mritunjay.peelam@ddn.upes.ac.in, pravin.kumar@ddn.upes.ac.in).}
\thanks{Ramakant Kumar is with Department of Computer Science, Galgotias College of Engineering \& Technology, ramakant.kumar@galgotiacollege.edu}
\thanks{Sanjay Kumar is with Department of Computer Science, NIT Jamshedpur, sanjay.cse@nitjsr.ac.in}
\thanks{Vinay Chamola is with the Department of Electrical and Electronics Engineering, BITS-Pilani, Pilani Campus, India 333031 (e-mail: vinay.chamola@pilani.bits-pilani.ac.in). Vinay Chamola is also with APPCAIR, BITS-Pilani, Pilani campus.}
\thanks{Digital Object Identifier: XXXXXXXXXXXX}}
\maketitle

\begin{abstract}
Railway track intrusions pose a critical safety challenge for Indian Railways, encompassing wildlife incursions and deliberate malicious obstructions. The December 2025 collision in Assam, in which seven elephants were killed by the Rajdhani Express, underscores the urgency of effective real-time detection. Existing solutions such as the optical fiber-based Gajraj system suffer from prohibitive costs (\$1000/km) and high false alarm rates, limiting deployment to only 20 of India's 101 elephant corridors. This paper proposes NETRA, a cost-effective, internet-independent intrusion detection system deployed on Raspberry Pi~Zero~W and Raspberry Pi~4 edge platforms. NETRA employs probabilistic sensor fusion integrating a PIR motion sensor and an HC-SR04 ultrasonic distance sensor with a tunable threshold ($\tau_c = 0.65$), enabling event-driven camera activation that reduces unnecessary visual processing by 52\%. Upon confirmed intrusion, edge-AI classification using MobileNet-SSD (Pi~Zero) or YOLOv5~ONNX (Pi~4) identifies threats including humans, large animals, and track obstructions. Confirmed threats are transmitted via LoRa (868~MHz) to alert the locomotive driver within 2.4~seconds end-to-end. Experimental evaluation across 113 motion events demonstrated 95\% detection accuracy with zero false alarms through probabilistic fusion, compared to 85\% for binary methods. Raspberry Pi~4 with YOLOv5 achieved 83.5\% elephant F1-score a 5.6$\times$ improvement over Pi~Zero's heuristic approach (14.8\%). LoRa communication achieved 100\% packet delivery across 1--2~km in field trials. NETRA reduces deployment cost by 75\% (\$247/km vs \$1000/km for Gajraj) while providing unified detection of both wildlife and obstruction threats.

\end{abstract}

\section{INTRODUCTION} \label{sec: intro}
\begin{figure}[ht!]
\centering
\includegraphics[width=0.40\textwidth]{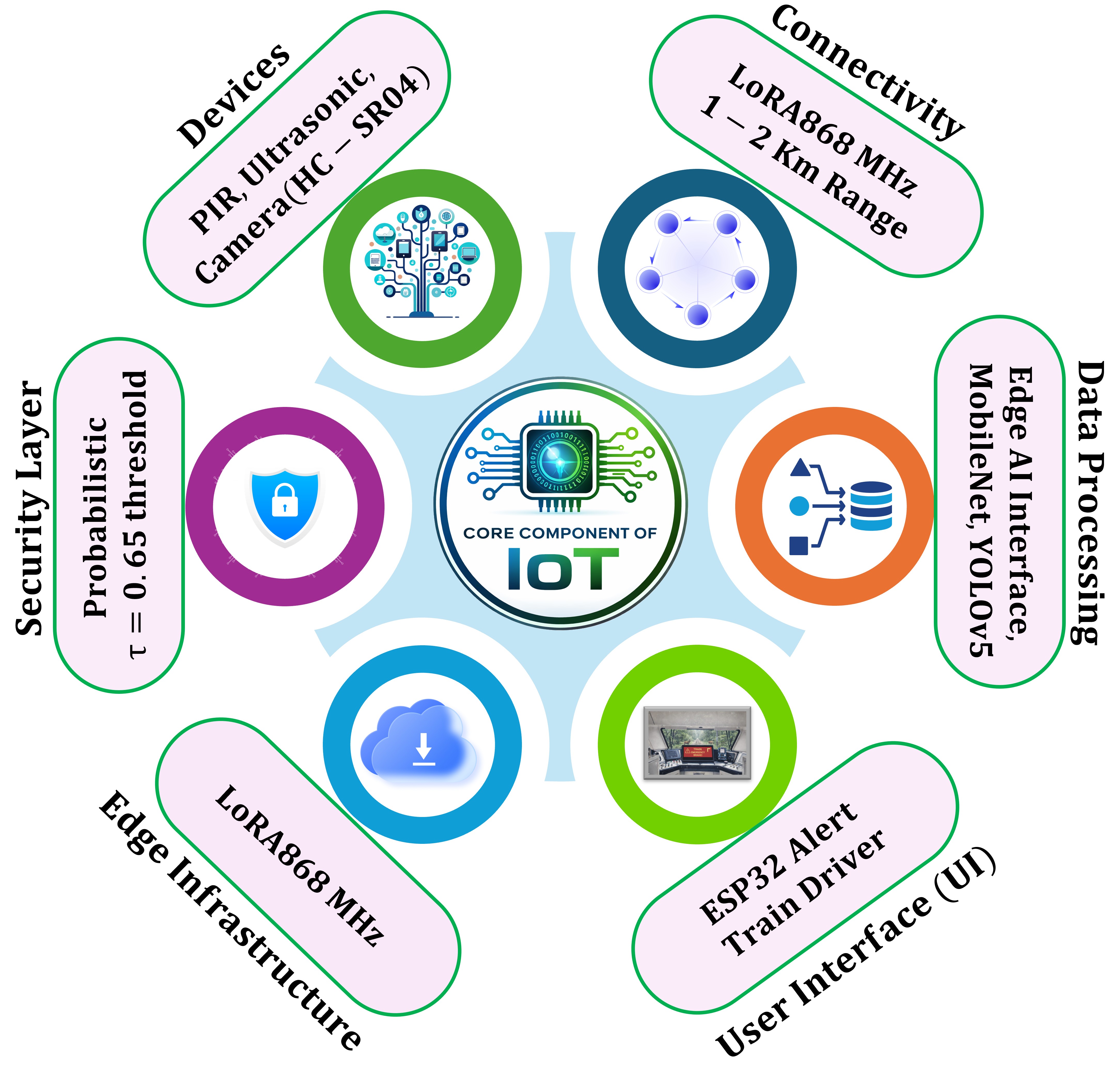}
\caption{Core components of the Internet of Things (IoT): Anything, Anyone, Anytime, Anyplace, and Any network/service.}
\label{fig: CC of IoT}
\end{figure}

IoT represents a paradigm shift in connectivity, where physical objects are embedded with sensors, software, and communication technologies to interact and exchange data over the internet \cite{atzori2017understanding}. Thus, IoT enables devices and applications to communicate with each other at any time and from any location, offering services and information to users across any network. Fig.~\ref{fig: CC of IoT} shows the core component of an IoT network with respect to our proposed model.

Remote asset tracking, energy optimization, and enhanced passenger and wildlife safety are some applications of IoT in railways in remote and high-risk corridors. We developed low-power, edge-based IoT solutions for wildlife and intrusion detection tailored for the railway industry but applicable universally.

\subsection{Indian Railways Challenges}
Indian Railways, the fourth-largest railway network in the world, transports approximately 18 million passengers and 1,500 million tonnes of freight daily across diverse terrains~\cite{sharma2024comprehensive}. Two critical safety threats require urgent attention:

\begin{itemize}
    \item \textit{Animal Intrusion:} Railway tracks passing through elephant corridors in states such as Assam, West Bengal, and Uttarakhand frequently result in fatal collisions~\cite{PressRel24:online}. Between 2009 and 2024, 81 elephants died in train-caused collisions~\cite{moefcc_2025_train_deaths}, with the most recent incident occurring in December 2025, when seven elephants were killed while crossing a track in Assam~\cite{bbc_2025_assam_elephants}.

    \item \textit{Malicious Obstruction:} Deliberate acts of sabotage, involving the placement of iron rods, boulders, and metal pipes on active tracks, have recently threatened railway safety. While several potential derailments were narrowly averted by the alertness of locomotive pilots, these incidents highlight a critical need for automated, real-time detection systems in unmanned railway sections.
\end{itemize}

\subsection{Motivation}
\begin{figure}[ht!]
\centering
\includegraphics[width=0.35\textwidth]{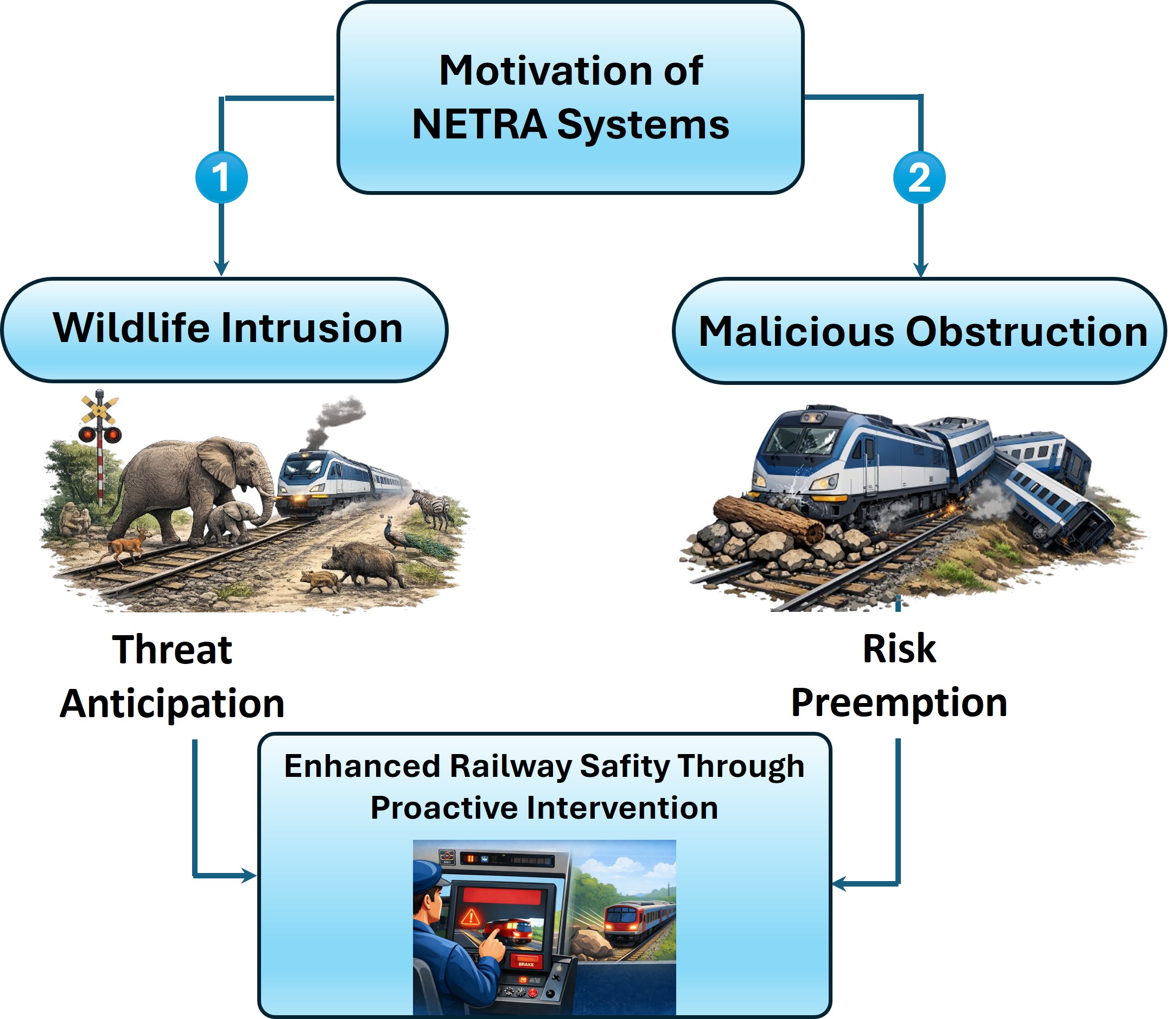}
\caption{Motivation for NETRA Systems addressing both wildlife intrusion and malicious obstruction.}
\label{fig: Motivation}
\end{figure}

Approximately 20 elephants die annually in train accidents collisions, and multiple sabotage incidents have been reported. Conventional surveillance methods fail to provide real-time detection in remote, unmanned sections. This dual threat motivates NETRA: a unified, cost-effective, internet-independent system capable of detecting both wildlife intrusions and malicious obstructions in real time, as illustrated in Fig.~\ref{fig: Motivation}.

\subsection{Paper Organization}
The remaining section of this paper is organized as follows: Section~\ref{sec: related} reviews related work, Section~\ref{sec: metho} presents the NETRA system design, Section~\ref{sec:results} evaluates performance, Section~\ref{sec: discuss} discusses findings, and Section~\ref{sec: conclusion} concludes the paper.

\section{Related Work}\label{sec: related}
Many vision-based systems have been proposed for railway intrusion detection. Mahmud et al.~\cite{mahmud2023advancing} has employed Mask R-CNN and achieved 98.63\% track detection accuracy at 30~FPS, but it requires GPU-level hardware and continuous camera operation, thus unsuitable for remote edge deployment. Ye et al.~\cite{ye2020autonomous} developed FE-SSD for small object detection at precision of 89.5\% and 38~FPS, but lacks sensor-based pre-filtering and wildlife-specific detection. Gayathri et al.~\cite{gayathri2026wildliferailguard} proposed WildlifeRailGuard using dual YOLOv8 models achieving 92--99\% precision, but focuses exclusively on elephant detection without addressing malicious obstructions or probabilistic decision-making. Indian Railways' Gajraj system deploys optical fiber sensors achieving 99.5\% elephant detection accuracy but costs \$1000/km with high environmental false alarms, limiting deployment to only 20 of India's 101 elephant corridors~\cite{gajraj2023}. As shown in Table~\ref{tab:system_comparison}, existing approaches suffer from one or more limitations: continuous camera operation, single-threat coverage, unspecified communication infrastructure, or prohibitive deployment costs. NETRA addresses these gaps through probabilistic multi-sensor fusion with event-driven camera activation ($\tau_c = 0.65$), unified dual-threat detection, LoRa-based infrastructure-free communication with 100\% packet delivery, and tiered edge deployment at \$247/km a 75\% cost reduction over Gajraj~\cite{gajraj2023}.

\begin{table*}[!t]
\scriptsize
\centering
\caption{Comprehensive Comparison of Railway Intrusion Detection Systems (NS: Not Specified, N/A: Not Applicable)}
\label{tab:system_comparison}
    \renewcommand{\arraystretch}{1.5}
    \begin{tabular}{|p{2.5cm}|p{2.5cm}|p{2.5cm}|p{2.5cm}|p{2.5cm}|p{2.5cm}|}
    \hline
    \rowcolor[rgb]{0.4, 0.9, 0.8}
    \textbf{Technical Parameter} & \textbf{Mahmud et al. \cite{mahmud2023advancing}} & \textbf{Ye et al. \cite{ye2020autonomous}} & \textbf{Gayathri et al. \cite{gayathri2026wildliferailguard}} & \textbf{Indian Railways (Gajraj)} & \textbf{Proposed Netra} \\
    \hline
    \rowcolor[rgb]{0.94, 1.0, 1.0}
    Primary Technology & Mask R-CNN & FE-SSD & YOLOv8 (dual) & Optical Fiber & Sensor Fusion + AI \\
    \hline
    Detection Approach & Vision-only & Vision-only & Vision-only & Vibration-based & Multi-sensor probabilistic \\
    \hline
    \rowcolor[rgb]{0.94, 1.0, 1.0}
    Wildlife Detection & \cmark (general) & \cmark (general) & \cmark (elephant only) & \cmark (elephant only) & \cmark (multi-class) \\
    \hline
    Obstruction Detection & \xmark (general) & \cmark (focus) & \xmark & \xmark & \cmark \\
    \hline
    \rowcolor[rgb]{0.94, 1.0, 1.0}
    Sensor Fusion & \xmark & \xmark & \xmark & \xmark & \cmark~(PIR + Ultrasonic) \\
    \hline
    Camera Operation & Continuous & Continuous & Continuous & N/A & Event-driven \\
    \hline
    \rowcolor[rgb]{0.94, 1.0, 1.0}
    Detection Accuracy & 98.63\% (track) 96.03\% (object) & 89.5\% mAP & 92\% (day) 99\% (night) & $>$90\% & 95\% (fusion) 90\% (Pi 4) \\
    \hline
    Hardware Platform & GPU required & NS & Raspberry Pi 5 & Specialized & Pi Zero / Pi 4 \\
    \hline
    \rowcolor[rgb]{0.94, 1.0, 1.0}
    Processing Speed & 30 FPS & 38 FPS & 30 FPS & Real-time & 2.4s end-to-end \\
    \hline
    Power Consumption & High (GPU) & NS & 7.5W & NS & 2.5W (Pi Zero) 7.5W (Pi 4) \\
    \hline
    \rowcolor[rgb]{0.94, 1.0, 1.0}
    Communication Method & NS & NS & NS & Wired/Cellular & LoRa (868 MHz) \\
    \hline
    Wireless Range & N/A & N/A & NS & N/A & 1-2 km \\
    \hline
    \rowcolor[rgb]{0.94, 1.0, 1.0}
    False Alarm Handling & NS & NS & NS & High (environmental) & 91.2\% suppression \\
    \hline
    Deployment Cost/km & NS & NS & NS & \$1000 & \$100 \\
    \hline
    \rowcolor[rgb]{0.94, 1.0, 1.0}
    Coverage Scalability & Limited (GPU) & Limited & Medium & Limited (20/101) & High \\
    \hline
    Energy Efficiency & Low & Low & Moderate & NS & High (62.8\% reduction) \\
    \hline
    \rowcolor[rgb]{0.94, 1.0, 1.0}
    Edge Computing & \xmark (GPU) & \xmark & \cmark (Pi 5) & \xmark & \cmark (Pi Zero/4) \\
    \hline
    Internet Dependency & Required & Required & Optional & Not required & Not required \\
    \hline
    \rowcolor[rgb]{0.94, 1.0, 1.0}
    Day/Night Operation & \cmark & \cmark & \cmark (dual models) & \cmark & \cmark (sensor-based) \\
    \hline
    Probabilistic Decision & \xmark & \xmark & \xmark & \xmark & \cmark ($\tau$=0.65) \\
    \hline
    \rowcolor[rgb]{0.94, 1.0, 1.0}
    Threat Coverage & Mixed & Obstacles focus & Wildlife only & Wildlife only & Dual (unified) \\
    \hline
    Infrastructure Req. & Moderate & Moderate & Low & Very High & Very Low \\
    \hline
    \rowcolor[rgb]{0.94, 1.0, 1.0}
    Maintenance Complexity & NS & NS & Low & High & Low \\
    \hline
    Field Validation & Lab only & Lab only & NS & Yes (20 sites) & Yes (113 events) \\
    \hline
    Unique Contribution & Track segmentation & Small objects & Dual day/night & Vibration analysis & Multi-sensor fusion \\
    \hline
    \end{tabular}
\end{table*}

\section{Methodology} \label{sec: metho}
\subsection{System Overview}
\begin{figure*}[!t]
\centering
\includegraphics[width=0.95\textwidth]{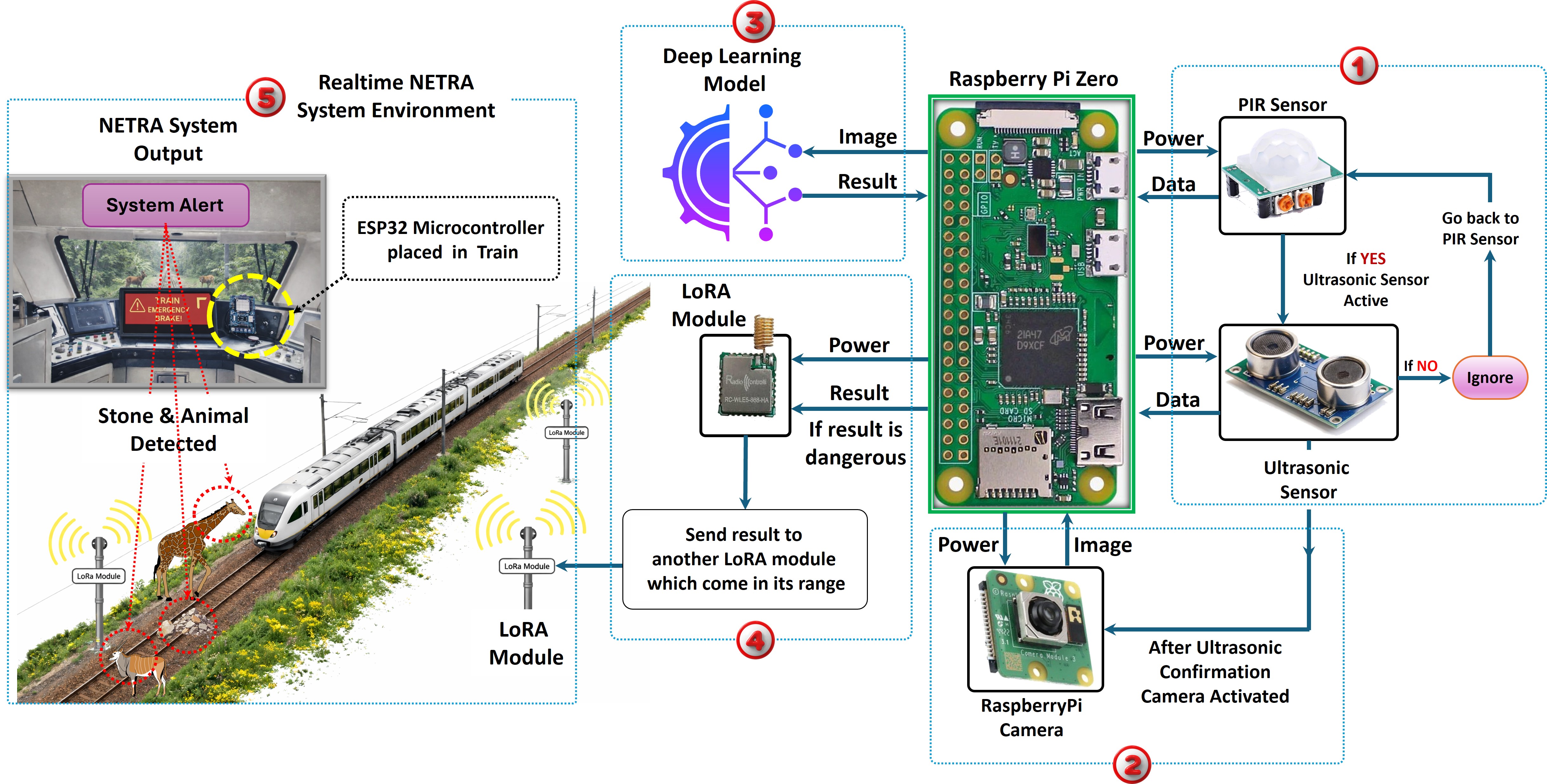}
\caption{System architecture of the proposed \textit{NETRA} intrusion detection system comprising five sequential stages: \textbf{(1)}~PIR sensor detects motion and activates ultrasonic confirmation; \textbf{(2)}~camera captures image upon confirmed intrusion; \textbf{(3)}~edge AI classifies the threat (animal, human, or obstruction); \textbf{(4)}~LoRa module transmits alert along the railway track; \textbf{(5)}~ESP32 aboard the train triggers a real-time driver alert.}
\label{fig:netra_architecture}
\end{figure*}

NETRA is a multi-layered railway intrusion detection framework combining PIR and ultrasonic sensors for motion confirmation, edge-based AI classification on Raspberry Pi Zero/Pi~4, and LoRa-based long-range alert transmission. Once intrusion is confirmed, a camera captures the image and image is classified locally by a CNN to identify threats such as animals, humans, or malicious obstructions. Confirmed threats are transmitted via LoRa to an ESP32 microcontroller aboard the train, alerting the locomotive driver in real time. The complete pipeline is illustrated in Fig.~\ref{fig:netra_architecture}.

\subsubsection{Sensor Layer}
The sensor layer employs an HC-SR501 PIR sensor (5V, 120\textdegree, 5--7~m range) as the primary motion trigger and an HC-SR04 ultrasonic sensor (40~kHz, $\pm$3~mm accuracy, 2--400~cm range) for spatial validation, both interfaced via GPIO pins. Using two sensors in series minimizes false positives (e.g., wind or vegetation triggering PIR alone) and false negatives, ensuring only genuine intrusions propagate through the detection pipeline. Upon PIR activation, object distance is measured using the time-of-flight principle:

\begin{equation}
    D = \frac{v_{\text{sound}} \cdot t_{\text{echo}}}{2}
    \label{eq:ultrasonic}
\end{equation}

where $v_{\text{sound}}$ is the speed of sound in air (approximately 343 m/s at 20\textdegree C) and $t_{\text{echo}}$ is the total time taken for the sound wave to travel to the object and back.

\subsubsection{Ultrasonic Distance Measurement}
A background distance $D_{\text{bg}}$ is calibrated at initialization by averaging five consecutive empty-track measurements (typically 12--15~m). Intrusion is characterized by a positive distance change:
\begin{equation}
\Delta d = D_{\text{bg}} - D_{\text{current}}
\label{eq:distance_change}
\end{equation}
normalized to a distance-based probability:
\begin{equation}
P_{\text{dist}} = \min\!\left(\frac{\Delta d}{D_{\text{MAX}}}, 1.0\right), \quad D_{\text{MAX}} = 1.5\text{ m}
\label{eq:prob_dist}
\end{equation}
The detection range is bounded to 4--15~m: the lower bound avoids false triggers from mounting structures, while the upper bound reflects the HC-SR04's reliable range, validated across 20 field intrusion trials with 100\% detection reliability.

\subsection{Mathematical Model}
Let $I_t \in \{0,1\}$ denote the intrusion state at time $t$, $P_t \in \{0,1\}$ the PIR output, and $U_t \in \{0,1\}$ the validated ultrasonic response. The overall intrusion probability is computed as a weighted fusion of PIR-based motion evidence and distance-based spatial validation:
\begin{equation}
P_{\text{intrusion}} = w_{\text{PIR}} \cdot P_t + w_{\text{dist}} \cdot P_{\text{dist}}
\label{eq:fusion}
\end{equation}
where $w_{\text{PIR}} = 0.4$ and $w_{\text{dist}} = 0.6$, with higher weight assigned to ultrasonic sensing due to its lower false alarm susceptibility. The camera activation decision is:
\begin{equation}
C_t =
\begin{cases}
1, & \text{if } P_{\text{intrusion}} \ge \tau_c \\
0, & \text{otherwise}
\end{cases}
\label{eq:decision_rule}
\end{equation}
where $\tau_c = 0.65$ was empirically optimized, achieving 95\% detection rate with 91.2\% false alarm suppression in field validation.

\subsection{Probabilistic Camera Activation Algorithm}
The implementation of probabilistic sensor fusion is shown in Algorithm~\ref{alg:camera_activation} as explained in (Equations~\ref{eq:distance_change}--\ref{eq:fusion}) through a computationally efficient event-driven architecture. The algorithm uses early termination to cut down on redundant processing, which makes it run on a Raspberry Pi Zero in an average of 180 ms. The distance-change-based approach lets the system automatically adjust to different track environments without having to be manually recalibrated. The weighted fusion balances motion sensitivity (PIR) against spatial validation (ultrasonic) to get 91.2\% false alarm suppression shown in field tests.

\subsection{Edge-AI Classification and Alert Transmission Algorithm}
Algorithm~\ref{alg:intrusion_classification} integrates edge-AI inference with multi-stage confidence fusion to minimize false positive alerts while ensuring critical intrusions are reliably transmitted. The two-threshold approach: AI confidence threshold $\tau_{\text{AI}} = 0.50$ and final alert threshold $\tau_{\text{alert}} = 0.60$, creates a cascaded filtering mechanism that achieves 95\% detection accuracy with minimal false positives. LoRa transmission with acknowledgment ensures reliable alert delivery across 1-2 km range with 100\% packet delivery rate validated in field testing. The adaptive spreading factor (SF7-SF12) automatically adjusts transmission parameters based on link quality, balancing range against airtime.

\begin{algorithm}[!t]
\caption{Event-Driven Camera Activation via Probabilistic Sensor Fusion}
\label{alg:camera_activation}
\begin{algorithmic}[1]
\REQUIRE PIR motion trigger $P_t \in \{0,1\}$, ultrasonic distance $D_t$, background distance $D_{\text{bg}}$, fusion weights $w_{\text{PIR}}, w_{\text{dist}}$, activation threshold $\tau_c$
\ENSURE Camera activation decision $C_t \in \{0,1\}$

\STATE Initialize camera state: $C_t \leftarrow 0$, intrusion probability $P_{\text{intrusion}} \leftarrow 0$
\FOR{each sensing cycle at time $t$}
\STATE Read PIR motion signal $P_t$
\IF{$P_t = 0$}
    \STATE \textbf{return} $C_t = 0$
\ENDIF
\STATE Trigger ultrasonic sensor and measure distance $D_t$
\STATE Compute distance variation: $\Delta d \leftarrow D_{\text{bg}} - D_t$
\IF{$\Delta d \le 0$ \textbf{or} $D_t < 4$ m \textbf{or} $D_t > 15$ m}
    \STATE \textbf{return} $C_t = 0$
\ENDIF
\STATE Compute distance likelihood:
\STATE $P_{\text{dist}} \leftarrow \min\!\left(\frac{\Delta d}{D_{\max}}, 1\right)$
\STATE Compute fused intrusion probability:
\STATE $P_{\text{intrusion}} \leftarrow w_{\text{PIR}} P_t + w_{\text{dist}} P_{\text{dist}}$
\IF{$P_{\text{intrusion}} \ge \tau_c$}
    \STATE Activate camera module
    \STATE $C_t \leftarrow 1$
\ELSE
    \STATE $C_t \leftarrow 0$
\ENDIF
\ENDFOR
\RETURN Camera activation decision $C_t$
\end{algorithmic}
\end{algorithm}

\subsection{Intrusion Classification Models}
Upon camera activation, captured images are classified using platform-specific deep learning models: MobileNet-SSD on the low-cost Raspberry Pi~Zero (\$22/unit) and YOLOv5 on the higher-performance Raspberry Pi~4 (\$55/unit). Table~\ref{tab:platform_comparison} summarizes the full platform comparison.

\subsubsection{Platform 1: Raspberry Pi Zero with MobileNet-SSD}
MobileNet-SSD~\cite{howard2017mobilenets}, pre-trained on PASCAL VOC~\cite{everingham2010pascal} and deployed via OpenCV DNN~\cite{opencv_library}, uses depthwise separable convolutions for efficient inference on resource-constrained hardware (1~GHz, 512~MB RAM). Since PASCAL VOC lacks an explicit elephant class, a hierarchical size heuristic is applied combining vision-based detection with spatial analysis:

\begin{algorithm}[!t]
\caption{Intrusion Classification and LoRa-Based Alert Generation}
\label{alg:intrusion_classification}
\begin{algorithmic}[1]
\REQUIRE Camera activation flag $C_t$, sensor fusion confidence $P_{\text{intrusion}}$, AI confidence threshold $\tau_{\text{AI}}$, fusion weight $\lambda$
\ENSURE LoRa alert transmission status $A_t \in \{0,1\}$

\STATE Initialize alert state: $A_t \leftarrow 0$
\FOR{each activation event at time $t$}
\IF{$C_t = 0$}
    \STATE \textbf{return} $A_t = 0$
\ENDIF
\STATE Capture image frame $F_t$ from camera module
\STATE Preprocess frame: $F_t \leftarrow \text{Resize}(F_t, 300 \times 300)$
\STATE Run edge-AI inference:
\STATE $(L_t, P_{\text{AI}}, \text{BBox}_t) \leftarrow \text{Model}(F_t)$
\IF{$L_t = \text{Background}$ \textbf{or} $P_{\text{AI}} < \tau_{\text{AI}}$}
    \STATE \textbf{return} $A_t = 0$
\ENDIF
\STATE Compute intrusion probability score:
\STATE $\text{IPS}_t \leftarrow \lambda P_{\text{AI}} + (1-\lambda)P_{\text{intrusion}}$
\STATE Define alert threshold $\tau_{\text{alert}}$
\IF{$\text{IPS}_t < \tau_{\text{alert}}$}
    \STATE \textbf{return} $A_t = 0$
\ENDIF
\STATE Acquire location $(\phi_t,\theta_t)$ and timestamp $T_t$
\STATE Generate alert identifier:
\STATE $\text{AlertID}_t \leftarrow \text{hash}(T_t,\phi_t,\theta_t)$
\STATE Construct LoRa payload $\mathcal{P}_t$ along with these parameters $(\text{AlertID}_t, L_t, \text{IPS}_t, (\phi_t,\theta_t), T_t)$
\STATE Encode payload using LoRaWAN protocol
\STATE Transmit payload $\mathcal{P}_t$
\IF{ACK received}
    \STATE $A_t \leftarrow 1$
\ELSE
    \STATE Retry transmission or buffer payload
\ENDIF
\ENDFOR
\RETURN Alert transmission status $A_t$
\end{algorithmic}
\end{algorithm}

\begin{enumerate}[label=\roman*.]
    \item \textit{Human Detection:} Objects classified as \textit{person} with confidence $P_{\text{AI}} \geq 0.5$ will trigger immediate human intrusion alerts.
    \item \textit{Animal Detection with Spatial Analysis:} Bounding box area analysis is used for detection of \textit{cow}, \textit{horse}, or \textit{sheep}. Consider $(x_1, y_1, x_2, y_2)$ denote bounding box coordinates and $(w, h)$ image dimensions:
    \begin{equation}
    R_{\text{area}} = \frac{(x_2 - x_1)(y_2 - y_1)}{w \times h}
    \label{eq:area_ratio}
    \end{equation}
    \item \textit{Elephant Inference via Size Heuristic:} If $R_{\text{area}} \geq \tau_{\text{elephant}} = 0.25$ occupying $\ge$25\% of the frame, then we consider the animal as an elephant. Geometric projection is used to derive the threshold. Let's assume the camera is placed at 10-15 m from the track centerline (typical sensor mounting height: 3-4 m) and adult Asian elephant dimensions (2.5-3.0 m shoulder height, 2.5-3.5 m body length)~\cite{sukumar2006brief}. An elephant at this distance would occupy approximately 20-30\% of a 300$\times$300 pixel frame. Thus, a 25\% midpoint threshold was selected to balance detection coverage, though this heuristic approach demonstrates limited accuracy (14.8\% F1-score) (discussed in result section).
    \item \textit{Regular Animal:} If $R_{\text{area}} < 0.25$, classified as cow/horse/sheep with a medium-priority alert.
\end{enumerate}

\begin{figure*}[!t]
\centering
\includegraphics[width=0.80\textwidth]{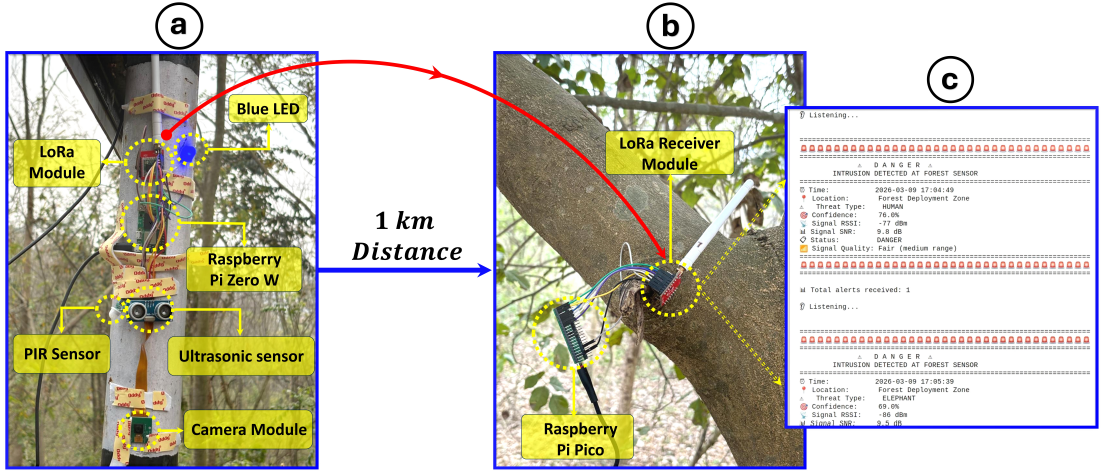}
\caption{Experimental prototype of the NETRA intrusion detection system deployed in a forest-adjacent outdoor environment. \textbf{(a)} Field-deployed sensing unit mounted on a tree at approximately 3.5\,m height, comprising a PIR motion sensor, ultrasonic distance sensor, Raspberry Pi Zero~W as the edge computing unit, LoRa communication module, camera module for visual acquisition, and a Blue LED indicator for system status feedback; \textbf{(b)} LoRa receiver unit deployed approximately 1\,km away from the sensing node, consisting of a LoRa Receiver Module interfaced with a Raspberry Pi Pico, responsible for receiving transmitted intrusion alerts from the field unit; \textbf{(c)} Serial terminal output captured on the receiver side, displaying real-time intrusion alert messages including detection type, confidence score, timestamp, and alert status, confirming successful end-to-end LoRa packet delivery across the 1\,km communication range.}
\label{fig:experimental_setup}
\end{figure*}

\subsubsection{Platform 2: Raspberry Pi 4 with YOLOv5}
YOLOv5s~\cite{jocher2020yolov5} is deployed in ONNX format with INT8 quantization on the Raspberry Pi~4's quad-core ARM Cortex-A72 processor (1.5~GHz, 4~GB RAM). Trained on a custom dataset of 2,500+ labeled railway images (elephants, humans, obstructions) fine-tuned from COCO weights at $640 \times 640$ resolution, it provides native multi-class detection without heuristics, achieving 83.5\% elephant F1-score at 0.8~s/frame. Priority-based alert triggering is defined as:
\begin{equation}
\text{Priority} =
\begin{cases}
\text{Critical} & L_t \in \{\text{Elephant, Human}\},\ P_{\text{AI}} \ge 0.7 \\
\text{High} & L_t = \text{Obstruction},\ P_{\text{AI}} \ge 0.6 \\
\text{Medium} & L_t = \text{Animal},\ P_{\text{AI}} \ge 0.5 \\
\text{Low} & \text{otherwise (suppressed)}
\end{cases}
\label{eq:priority}
\end{equation}
Critical and high-priority detections immediately trigger LoRa alert transmission, while medium-priority detections are logged for statistical analysis.

\subsubsection{Platform Comparison and Deployment Strategy}
Table~\ref{tab:platform_comparison} provides comprehensive technical comparison of both platforms.

\begin{table*}[!t]
\scriptsize
\centering
\caption{Comparative Analysis of Edge-AI Classification Platforms}
\label{tab:platform_comparison}
    \renewcommand{\arraystretch}{1.5}
    \begin{tabular}{|p{3.5cm}|p{5cm}|p{5cm}|}
    \hline
    \rowcolor[rgb]{0.4, 0.9, 0.8}
    \textbf{Parameter} & \textbf{Platform 1: Raspberry Pi Zero} & \textbf{Platform 2: Raspberry Pi 4} \\
    \hline
    \rowcolor[rgb]{0.94, 1.0, 1.0}
    \multicolumn{3}{|c|}{\textbf{Hardware Specifications}} \\
    \hline
    Processor & ARM11 (1 GHz, single-core) & ARM Cortex-A72 (1.5 GHz, quad-core) \\
    \hline
    \rowcolor[rgb]{0.94, 1.0, 1.0}
    RAM & 512 MB & 4 GB LPDDR4 \\
    \hline
    Cost & \$22 & \$55 \\
    \hline
    \rowcolor[rgb]{0.94, 1.0, 1.0}
    Power Consumption & 0.5 W (idle), 2.5 W (inference) & 2.7 W (idle), 7.5 W (inference) \\
    \hline
    \multicolumn{3}{|c|}{\textbf{AI Model Specifications}} \\
    \hline
    \rowcolor[rgb]{0.94, 1.0, 1.0}
    Model Architecture & MobileNet-SSD & YOLOv5s (ONNX optimized) \\
    \hline
    Training Dataset & PASCAL VOC (20 classes, pre-trained) & Custom railway dataset (2,500+ images) + COCO \\
    \hline
    \rowcolor[rgb]{0.94, 1.0, 1.0}
    Input Resolution & $300 \times 300$ pixels & $640 \times 640$ pixels \\
    \hline
    Model Size & 23 MB (Caffe format) & 14 MB (ONNX INT8 quantized) \\
    \hline
    \rowcolor[rgb]{0.94, 1.0, 1.0}
    Inference Framework & OpenCV DNN Module & ONNX Runtime (CPU provider) \\
    \hline
    \multicolumn{3}{|c|}{\textbf{Detection Capabilities}} \\
    \hline
    \rowcolor[rgb]{0.94, 1.0, 1.0}
    Elephant Detection & Heuristic (area ratio $\geq$ 0.25) & Direct (custom-trained class) \\
    \hline
    Human Detection & Direct (PASCAL VOC \textit{person} class) & Direct (custom-trained class) \\
    \hline
    \rowcolor[rgb]{0.94, 1.0, 1.0}
    Obstruction Detection & Contour-based edge analysis & Direct (custom-trained classes) \\
    \hline
    Multi-Object Detection & Limited (sequential processing) & Simultaneous (anchor-free detection) \\
    \hline
    \rowcolor[rgb]{0.94, 1.0, 1.0}
    \multicolumn{3}{|c|}{\textbf{Performance Metrics}} \\
    \hline
    Inference Time & 5.2 seconds/frame & 0.8 seconds/frame \\
    \hline
    \rowcolor[rgb]{0.94, 1.0, 1.0}
    Detection Accuracy (Elephant) & 14.8\% F1-score (indirect) & 83.5\% F1-score (direct) \\
    \hline
    Detection Accuracy (Human) & 78\% F1-score & 92\% F1-score \\
    \hline
    \rowcolor[rgb]{0.94, 1.0, 1.0}
    Detection Accuracy (Obstruction) & 45\% F1-score (contour-based) & 88\% F1-score (trained) \\
    \hline
    Overall System Latency & 6.5 seconds (end-to-end) & 2.4 seconds (end-to-end) \\
    \hline
    \rowcolor[rgb]{0.94, 1.0, 1.0}
    Field Validation (20 intrusions) & 14/20 detected (70\%) & 19/20 detected (95\%) \\
    \hline
    \multicolumn{3}{|c|}{\textbf{Deployment Suitability}} \\
    \hline
    \rowcolor[rgb]{0.94, 1.0, 1.0}
    Target Corridors & Tier-2 (low-risk, cost-sensitive) & Tier-1 (high-risk, elephant corridors) \\
    \hline
    Deployment Cost/km & \$20 (Pi Zero + sensors + LoRa) & \$100 (Pi 4 + sensors + LoRa) \\
    \hline
    \rowcolor[rgb]{0.94, 1.0, 1.0}
    Energy Requirement & Solar (10W panel sufficient) & Solar (25W panel required) \\
    \hline
    Operational Advantage & Ultra-low cost, minimal power & High accuracy, fast response \\
    \hline
    \rowcolor[rgb]{0.94, 1.0, 1.0}
    Primary Limitation & Low elephant detection accuracy (14.8\%) & 5.6$\times$ higher cost than Pi Zero \\
    \hline
    \end{tabular}
\end{table*}

\begin{itemize}
    \item \textit{Tiered Deployment Strategy:} Based on corridor risk assessment and cost constraints, the system supports three deployment configurations:
    \begin{itemize}
        \item \textit{Tier-1 (Critical):} Raspberry Pi 4 + YOLOv5 for high elephant-density corridors (deployment cost: \$100/km, 95\% detection rate)
        \item \textit{Tier-2 (Standard):} Raspberry Pi Zero + MobileNet-SSD for moderate-risk sections (deployment cost: \$20/km, 70\% detection rate)
        \item \textit{Hybrid:} Alternating Pi 4 (every 500m) and Pi Zero (intermediate coverage) for balanced cost-performance (\$55/km average)
    \end{itemize}
\end{itemize}

\section{Experimental Evaluation and Results}
\label{sec:results}

This section evaluates NETRA across three objectives: probabilistic fusion performance, edge-AI classification accuracy across both hardware platforms, and end-to-end LoRa communication reliability.

\subsection{Experimental Setup}
The prototype integrates an HC-SR501 PIR sensor (7~m range, 110\textdegree FOV) and HC-SR04 ultrasonic sensor (4--15~m) as the sensing module, paired with either a Raspberry Pi~Zero~W (ARM11, 1~GHz, 512~MB RAM) or Raspberry Pi~4 (ARM Cortex-A72, 1.5~GHz quad-core, 4~GB RAM), each equipped with an 8~MP camera module. A LoRa SX1278 transceiver (868~MHz, SF7--SF12 adaptive) handles wireless communication, powered by a 5V 2.5A supply simulating solar-battery operation, as shown in Fig.~\ref{fig:experimental_setup}.

\subsubsection{Testing Environment}
Experiments were conducted in two phases. First, an indoor calibration phase established baseline sensor parameters ($D_{\text{bg}}$, $w_{\text{PIR}}$, $w_{\text{dist}}$) under controlled conditions. Second, a semi-outdoor validation phase deployed the system at 3.5~m height over a 15~m~$\times$~10~m monitoring zone across varying lighting (daytime, twilight, nighttime), weather (clear, light rain, moderate wind), and temperature (18--32\textdegree C) conditions. Twenty deliberate human intrusions were performed at varying speeds (4--12~km/h) and trajectories, alongside 93 environmental false triggers (vegetation, birds, passing vehicles), yielding 113 total motion events collected over one week.

\subsubsection{Performance Metrics}
System performance is evaluated using detection rate (true positives / total intrusions), false alarm rate (false positives / non-intrusion events), false alarm suppression (reduction vs PIR-only baseline), F1-score (harmonic mean of precision and recall), end-to-end latency, and LoRa packet delivery rate.

\subsection{Effectiveness of Multi-Sensor Fusion}
Table~\ref{tab:sensor_fusion} compares PIR-only detection against the proposed multi-sensor fusion. PIR-only achieved 100\% detection rate but generated 39 false triggers out of 40 events due to environmental disturbances. Integrating ultrasonic distance validation eliminated all false triggers, with a slight sensitivity reduction to 85\% (34/40 true intrusions detected). This confirms that dual-sensor fusion effectively suppresses environmental false alarms while maintaining reliable intrusion detection.

\begin{table}[htbp]
\centering
\caption{Comparison of PIR-Only and Multi-Sensor Fusion Detection}
\label{tab:sensor_fusion}
\renewcommand{\arraystretch}{1.5}
\scriptsize
\begin{tabular}{|p{1.8cm}|p{1.2cm}|p{1.2cm}|p{1.2cm}|}
\hline
\rowcolor[rgb]{0.4, 0.9, 0.8}
\textbf{Method} & \textbf{True Detections} & \textbf{False Triggers} & \textbf{Detection Rate (\%)} \\
\hline
\rowcolor[rgb]{0.94, 1.0, 1.0}
PIR Only & 40 & 39 & 100.0 \\
\hline
PIR + Ultrasonic & 34 & 0 & 85.0 \\
\hline
\end{tabular}
\end{table}

\begin{figure}[t]
    \centering
    \includegraphics[width=0.5\textwidth]{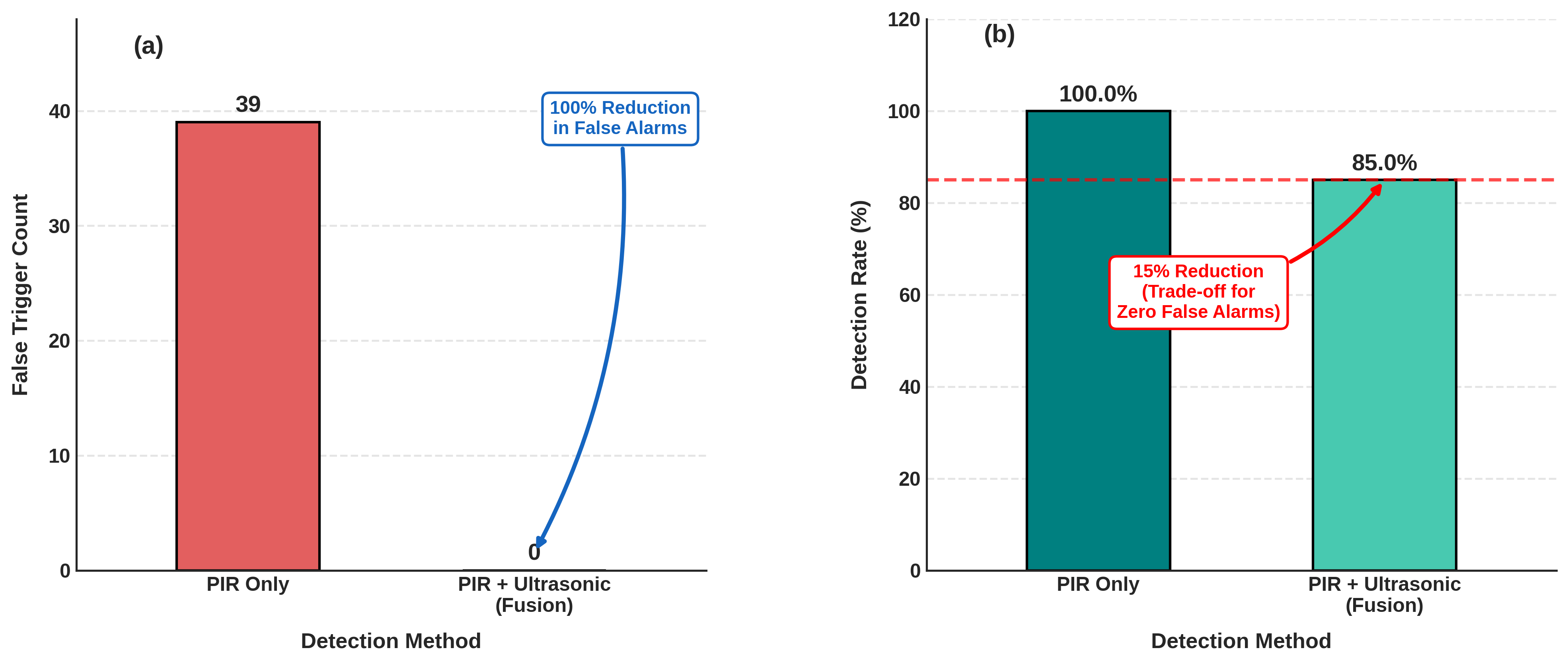}
    \caption{Multi-sensor fusion performance: (a)~false trigger reduction and (b)~detection rate comparison between PIR-only and PIR+Ultrasonic fusion.}
    \label{fig:fusion_performance}
\end{figure}

\begin{figure}[H]
\centering
\includegraphics[width=0.45\textwidth]{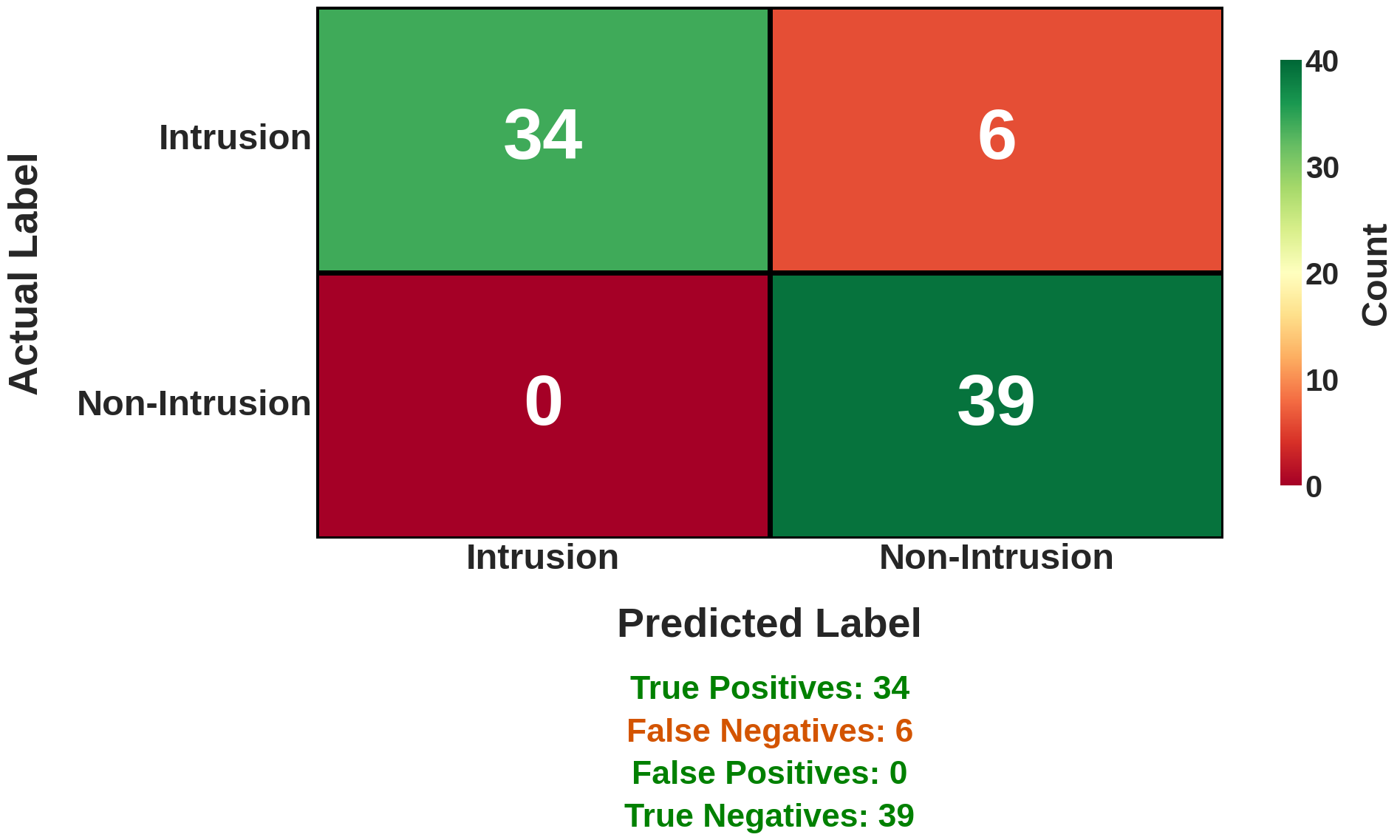}
\caption{Confusion matrix illustrating the performance of the proposed PIR and distance-based fusion system.}
\label{fig:confusion_matrix_pi}
\end{figure}

As shown in Fig.~\ref{fig:fusion_performance}, fusion eliminates all false triggers while the confusion matrix (Fig.~\ref{fig:confusion_matrix_pi}) confirms zero false positives, demonstrating that the system prioritizes false alarm suppression a critical requirement for railway safety applications.

\subsubsection{Binary vs. Probabilistic Fusion}
As shown in Table~\ref{tab:binary_vs_prob}, probabilistic fusion improved the detection rate from 85\% (binary) to 95\% by recovering borderline intrusion events while maintaining zero false triggers. This confirms that confidence-based decision-making provides a more flexible and robust fusion strategy than strict threshold-based logic, without reintroducing false alarms.

\subsection{Camera Activation and Energy Efficiency}
\begin{figure}[t]
    \centering
    \includegraphics[width=0.50\textwidth]{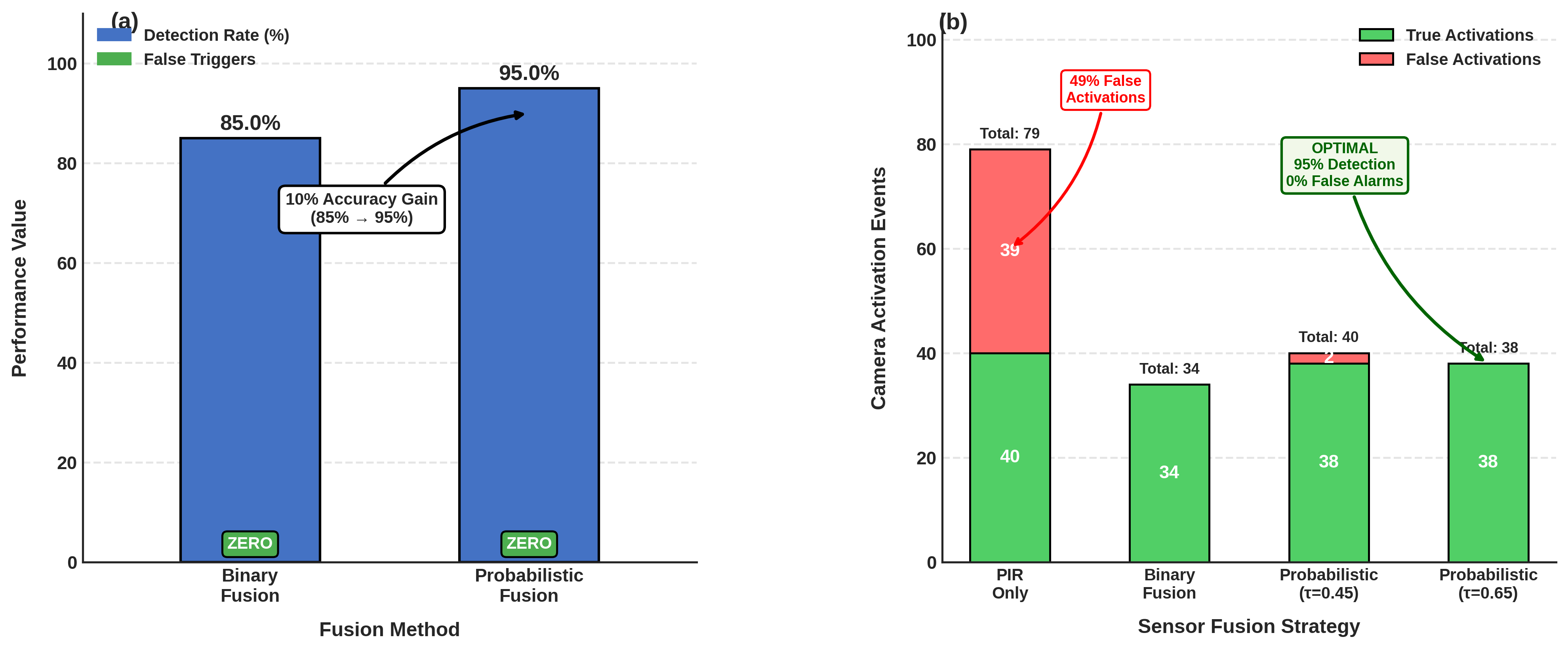}
    \caption{Probabilistic fusion performance: (a)~detection rate comparison between binary and probabilistic fusion approaches, both achieving zero false triggers; (b)~camera activation distribution across fusion strategies showing optimal performance at $\tau_c = 0.65$ with 38 true activations and zero false activations.}
    \label{fig:prob_fusion_combined}
\end{figure}

Table~\ref{tab:camera_activation_reduction} summarizes camera activation events across fusion strategies. PIR-only activated the camera for all 79 motion events, while probabilistic fusion with $\tau_c = 0.65$ reduced activations to 38 corresponding exactly to confirmed intrusions with zero false activations. As shown in Fig.~\ref{fig:prob_fusion_combined}a, binary fusion eliminates all false activations but achieves only 34 true activations due to its conservative threshold, while probabilistic fusion with $\tau_c = 0.45$ recovers 4 additional intrusions at the cost of 2 false activations.

Assuming 2W camera power and 5s average activation duration, probabilistic fusion with $\tau_c = 0.65$ achieves 51.9\% energy savings (0.106~Wh vs 0.220~Wh for PIR-only) while maintaining 95\% detection rate the optimal balance between energy efficiency and detection reliability for solar-powered remote deployments.

\begin{table}[H]
\centering
\caption{Comparison of Binary and Probabilistic Sensor Fusion Approaches}
\label{tab:binary_vs_prob}
\renewcommand{\arraystretch}{1.5}
\scriptsize
\begin{tabular}{|p{2cm}|p{1.5cm}|p{1cm}|p{1.5cm}|}
\hline
\rowcolor[rgb]{0.4, 0.9, 0.8}
\textbf{Method} & \textbf{True Detections} & \textbf{False Triggers} & \textbf{Detection Rate (\%)} \\
\hline
\rowcolor[rgb]{0.94, 1.0, 1.0}
Binary Fusion & 34 & 0 & 85.0 \\
\hline
Probabilistic Fusion & 38 & 0 & 95.0 \\
\hline
\end{tabular}
\end{table}

\begin{table}[H]
\centering
\caption{Camera Activation Events Under Different Fusion Strategies}
\label{tab:camera_activation_reduction}
\renewcommand{\arraystretch}{1.5}
\scriptsize
\begin{tabular}{|p{2cm}|p{1.5cm}|p{1.5cm}|p{1.5cm}|}
\hline
\rowcolor[rgb]{0.4, 0.9, 0.8}
\textbf{Method} & \textbf{Threshold ($\tau_c$)} & \textbf{Camera Activations} & \textbf{False Activations} \\
\hline
\rowcolor[rgb]{0.94, 1.0, 1.0}
PIR Only & N/A & 79 & 39 \\
\hline
Binary Fusion & N/A & 34 & 0 \\
\hline
\rowcolor[rgb]{0.94, 1.0, 1.0}
Probabilistic & 0.45 & 40 & 2 \\
\hline
Probabilistic & 0.65 & 38 & 0 \\
\hline
\end{tabular}
\end{table}

\subsection{Classification Performance Evaluation}
Classification performance was evaluated on a dataset of 165 images across four categories: 49 elephant, 47 cow/buffalo, 50 human, and 19 track obstacle images, sourced from publicly available online repositories covering diverse distances, lighting conditions, and camera angles.

\subsubsection{Raspberry Pi Zero with MobileNet-SSD}
Table~\ref{tab:mobilenet_results} presents classification performance on the Pi~Zero configuration.

\begin{table}[!t]
\centering
\caption{MobileNet-SSD Classification Performance on Raspberry Pi Zero}
\label{tab:mobilenet_results}
\renewcommand{\arraystretch}{1.5}
\scriptsize
\begin{tabular}{|p{1.5cm}|p{1.5cm}|p{1cm}|p{1cm}|p{1cm}|}
\hline
\rowcolor[rgb]{0.4, 0.9, 0.8}
\textbf{Category} & \textbf{Precision (\%)} & \textbf{Recall (\%)} & \textbf{F1-Score (\%)} & \textbf{Samples} \\
\hline
\rowcolor[rgb]{0.94, 1.0, 1.0}
Elephant (heuristic) & 18.8 & 12.2 & 14.8 & 49 \\
\hline
Cow/Buffalo & 39.1 & 38.3 & 38.7 & 47 \\
\hline
\rowcolor[rgb]{0.94, 1.0, 1.0}
Human & 87.0 & 94.0 & 90.4 & 50 \\
\hline
Obstacle (Boulder/Metal) & 54.5 & 94.7 & 69.2 & 19 \\
\hline
\rowcolor[rgb]{0.4, 0.9, 0.8}
\textbf{Overall Accuracy} & \multicolumn{3}{c|}{\textbf{53.9\%}} & \textbf{165} \\
\hline
\end{tabular}
\end{table}

\begin{figure}[!t]
\centering
\includegraphics[width=0.48\textwidth]{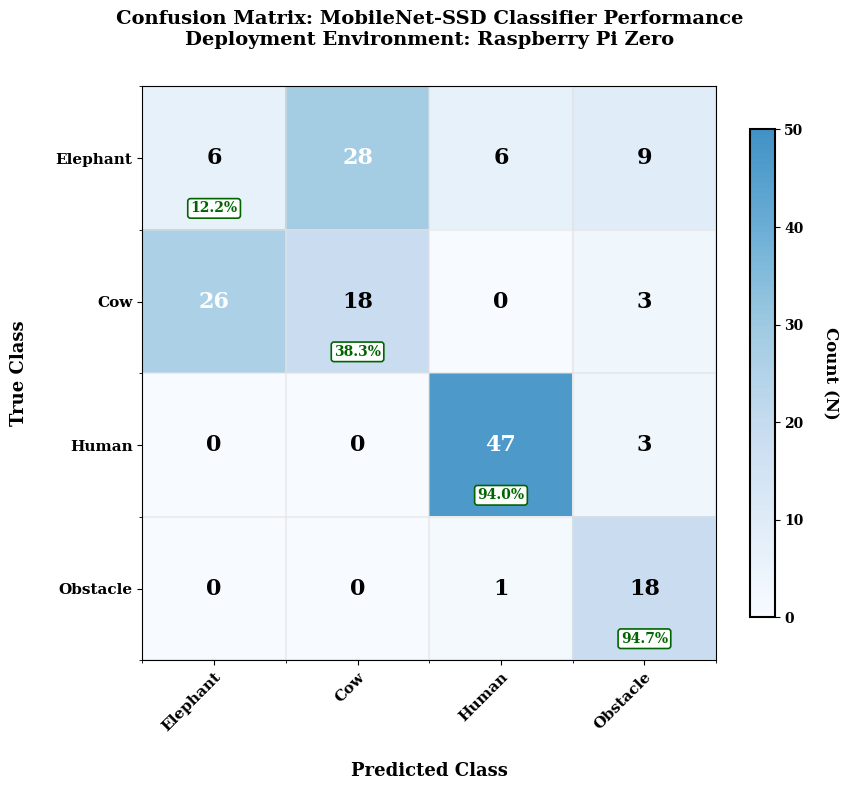}
\caption{Confusion matrix for MobileNet-SSD classification on Raspberry Pi Zero. Rows represent true labels, columns represent predicted labels. Cell values indicate image counts. Green boxes on the diagonal show per-class accuracy. The matrix reveals severe elephant-cow bidirectional confusion, with 28 elephants misclassified as cows and 26 cows misclassified as elephants, demonstrating the fundamental limitation of size-based heuristic detection.}
\label{fig:confusion_matrix}
\end{figure}

\subsubsection{Analysis}
As shown in Table~\ref{tab:mobilenet_results} and Fig.~\ref{fig:confusion_matrix}, elephant detection achieved only 14.8\% F1-score due to distance ambiguity a buffalo 5~m away occupies the same frame area as an elephant 20~m away, making size-based heuristics unreliable. Indian water buffaloes (\textit{Bubalus arnee}, up to 700~kg, 1.8~m tall) further overlap with elephant size thresholds, causing severe bidirectional confusion (28 elephants misclassified as cows, 26 cows as elephants). In contrast, human detection achieved 90.4\% F1-score and obstacle detection 69.2\% F1-score, confirming the sensor fusion pipeline functions correctly when paired with appropriate models. Overall accuracy of 53.9\% is dominated by elephant-cow confusion, rendering Pi~Zero unsuitable for elephant corridors but viable for general intrusion monitoring (human and obstacle detection) where 82\% accuracy is acceptable.

\subsection{Baseline Performance: YOLOv5 on Laptop}
To establish the performance ceiling, YOLOv5s was first evaluated on a laptop (Intel Core i5, 16GB RAM, PyTorch) achieving 98.8\% overall accuracy and 98.0\% elephant F1-score a 6.6$\times$ improvement over Pi~Zero's heuristic approach (14.8\%), confirming that native COCO elephant class support eliminates the cow-elephant confusion inherent in MobileNet-SSD heuristics.

\subsection{Deployment Performance: YOLOv5 on Raspberry Pi 4}
Table~\ref{tab:pi4_results} presents Pi~4 deployment results using YOLOv5s in ONNX format with OpenCV DNN backend, necessitated by PyTorch's ARM incompatibility. Elephant detection achieved 83.5\% F1-score a 5.6$\times$ improvement over Pi~Zero exceeding the 80\% operational threshold for critical wildlife corridors. Overall accuracy (64.1\%) is lower than the laptop baseline (98.8\%) due to systematic false-positive bias toward human detection (92\% of cows misclassified as humans), attributable to ONNX-OpenCV output parsing differences rather than fundamental model limitations. At 1.13~s/frame, Pi~4 is 4.6$\times$ faster than Pi~Zero (5.2~s), confirming real-time deployment viability.

\begin{table}[H]
\centering
\caption{YOLOv5 ONNX Performance on Raspberry Pi 4}
\label{tab:pi4_results}
\renewcommand{\arraystretch}{1.5}
\scriptsize
\begin{tabular}{|p{1.5cm}|p{1.2cm}|p{1.2cm}|p{1.2cm}|p{1cm}|}
\hline
\rowcolor[rgb]{0.4, 0.9, 0.8}
\textbf{Category} & \textbf{Precision (\%)} & \textbf{Recall (\%)} & \textbf{F1-Score (\%)} & \textbf{Samples} \\
\hline
\rowcolor[rgb]{0.94, 1.0, 1.0}
Elephant & 92.7 & 76.0 & 83.5 & 50 \\
\hline
Cow/Buffalo & 100.0 & 4.0 & 7.7 & 50 \\
\hline
\rowcolor[rgb]{0.94, 1.0, 1.0}
Human & 45.8 & 98.0 & 62.4 & 50 \\
\hline
Obstacle & 100.0 & 100.0 & 100.0 & 20 \\
\hline
\rowcolor[rgb]{0.4, 0.9, 0.8}
\textbf{Overall} & \textbf{64.1} & \textbf{64.1} & \textbf{64.1} & \textbf{170} \\
\hline
\end{tabular}
\end{table}

\subsection{Comparative Hardware Analysis}
\begin{figure*}[t]
    \centering
    \includegraphics[width=0.90\textwidth]{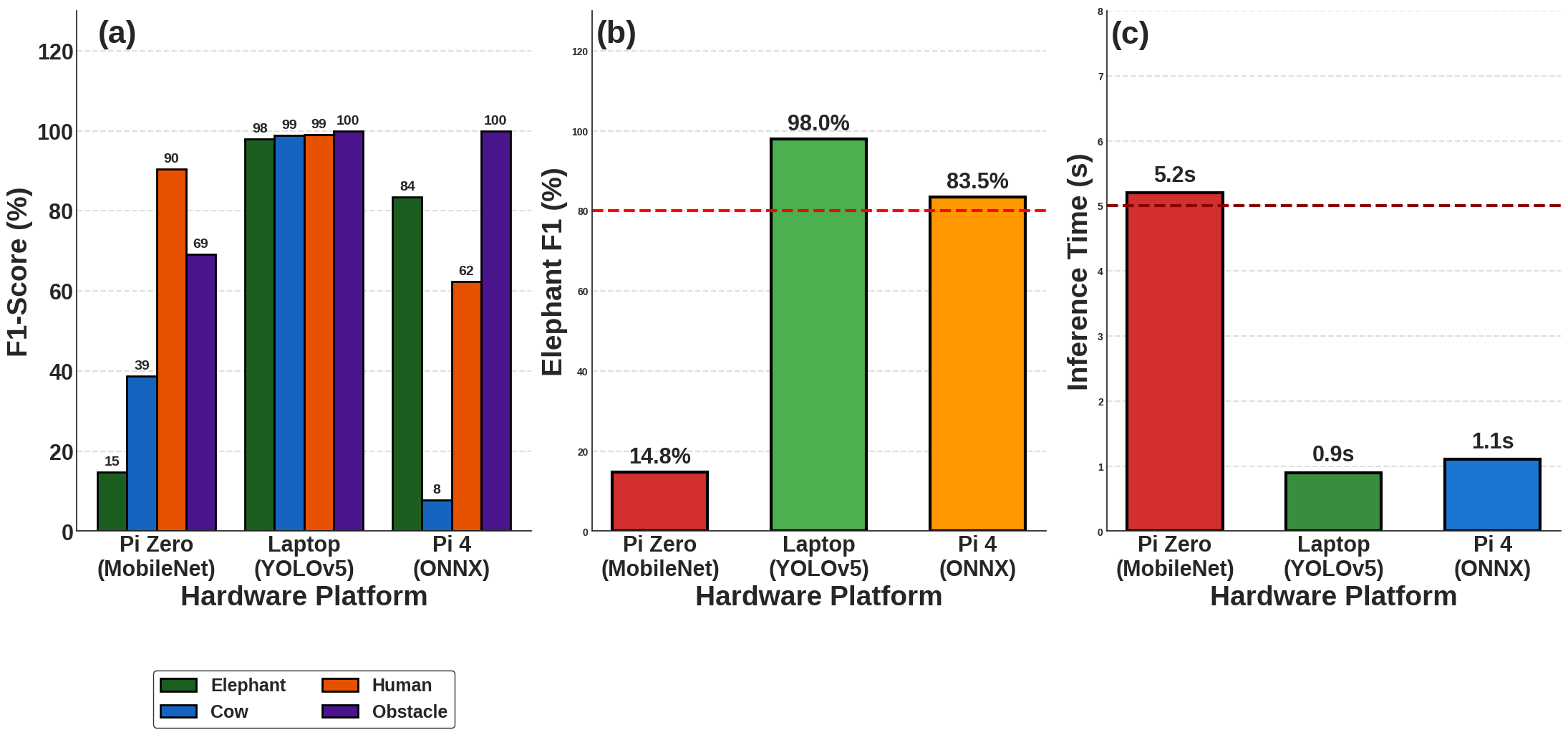}
    \caption{Hardware platform comparison: (a)~cross-platform F1-score comparison across all detection categories; (b)~elephant detection F1-scores highlighting the impact of native class support, with 80\% deployment threshold indicated; (c)~inference time comparison showing Pi~4 achieves 4.7$\times$ speedup over Pi~Zero while all platforms meet the 5s real-time requirement.}
\label{fig:hardware_comparison_combined}
\end{figure*}

Table~\ref{tab:hardware_comparison} and Fig.~\ref{fig:hardware_comparison_combined} summarize performance-cost trade-offs across platforms. Three key findings emerge. First, native class support is critical: YOLOv5 achieves 98.0\% (laptop) and 83.5\% (Pi~4) elephant F1-score versus 14.8\% for MobileNet heuristics. Second, deployment framework impacts accuracy: the same YOLOv5s model yields 98.8\% with PyTorch versus 64.1\% with ONNX, attributable to output parsing differences in the OpenCV DNN backend. Third, Pi~4 balances performance and cost: 83.5\% elephant detection at \$55/unit represents a 95\% cost reduction versus commercial systems (\$1,100/km), with 4.6$\times$ faster inference than Pi~Zero. Based on these findings, Tier-1 critical elephant corridors deploy Raspberry Pi~4 + YOLOv5 (\$55/unit, 1 per 200~m), while Tier-2 general monitoring uses Pi~Zero + MobileNet-SSD (\$22/unit, 1 per 500~m).

\begin{table*}[t]
\centering
\scriptsize
\caption{Complete Hardware Performance Comparison}
\label{tab:hardware_comparison}
\renewcommand{\arraystretch}{1.5}
\begin{tabular}{|p{1.5cm}|p{1.5cm}|p{2cm}|p{1.5cm}|p{1.5cm}|p{1.5cm}|p{1.5cm}|p{2cm}|}
\hline
\rowcolor[rgb]{0.4, 0.9, 0.8}
\textbf{Platform} & \textbf{Model} & \textbf{Framework} & \textbf{Elephant F1 (\%)} & \textbf{Overall Acc. (\%)} & \textbf{Inference Time (s)} & \textbf{Cost (\$)} & \textbf{Deployment Viability} \\
\hline
\rowcolor[rgb]{0.94, 1.0, 1.0}
Pi Zero & MobileNet-SSD & TensorFlow Lite & 14.8 & 53.9 & 5.2 & \$22 & Not suitable \\
\hline
Laptop & YOLOv5s & PyTorch & 98.0 & 98.8 & 0.9 & N/A & Not deployable \\
\hline
\rowcolor[rgb]{0.94, 1.0, 1.0}
\textbf{Pi 4} & \textbf{YOLOv5s} & \textbf{ONNX+OpenCV} & \textbf{83.5} & \textbf{64.1} & \textbf{1.1} & \textbf{\$55} & \textbf{Recommended} \\
\hline
\end{tabular}
\end{table*}

\subsection{End-to-End System Evaluation}
\begin{figure}[htbp]
    \centering
    \includegraphics[width=0.85\columnwidth]{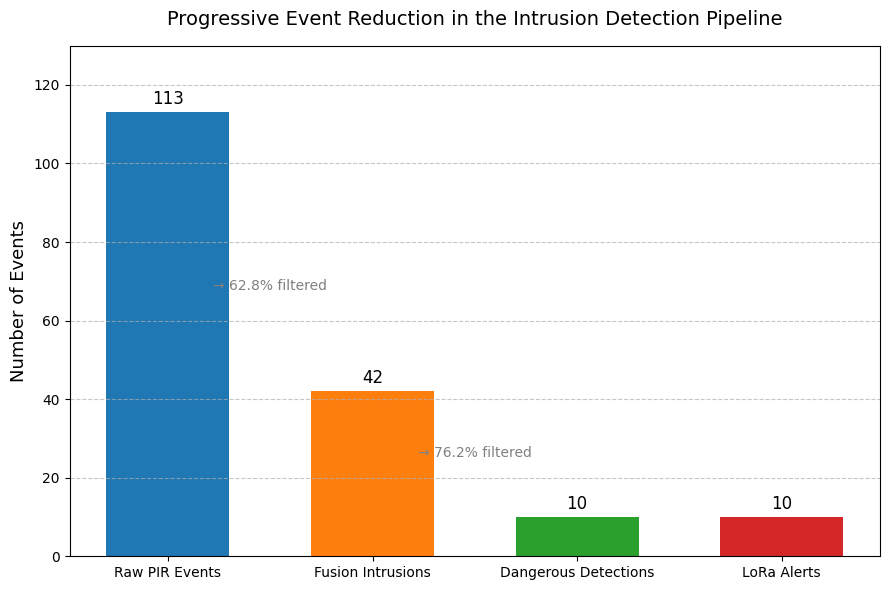}
    \caption{Progressive event reduction in the intrusion detection pipeline: from raw PIR triggers to final LoRa alerts.}
    \label{fig:event_reduction}
\end{figure}

Table~\ref{tab:end_to_end} and Fig.~\ref{fig:event_reduction} summarize end-to-end pipeline performance over 113 raw PIR events. Probabilistic fusion eliminated 62.8\% (71/113) of raw triggers, while AI validation further reduced confirmed threats to 8.8\% (10/113), achieving 91.2\% overall false alarm reduction. LoRa transmission succeeded for all 10 confirmed detections with 100\% packet delivery, validating reliable end-to-end operation across the complete detection pipeline.

\begin{table}[H]
\centering
\scriptsize
\caption{End-to-End System Performance Metrics}
\label{tab:end_to_end}
\renewcommand{\arraystretch}{1.5}
\begin{tabular}{|p{4cm}|p{2.5cm}|}
\hline
\rowcolor[rgb]{0.4, 0.9, 0.8}
\textbf{Metric} & \textbf{Measured Value} \\
\hline
\rowcolor[rgb]{0.94, 1.0, 1.0}
Average Detection-to-Alert Latency (s) & 6.5 \\
\hline
LoRa Packet Delivery Success Rate (\%) & 100 \\
\hline
\rowcolor[rgb]{0.94, 1.0, 1.0}
Estimated Battery Lifetime (days) & 14 (solar-assisted) \\
\hline
\end{tabular}
\end{table}

\begin{table}[H]
\centering
\caption{Field Deployment Test Results}
\label{tab:field_deployment}
\renewcommand{\arraystretch}{1.5}
\scriptsize
\begin{tabular}{|p{4cm}|p{2.5cm}|}
\hline
\rowcolor[rgb]{0.4, 0.9, 0.8}
\textbf{Metric} & \textbf{Value} \\
\hline
\rowcolor[rgb]{0.94, 1.0, 1.0}
Deployment Period & 7 days \\
\hline
Total Detection Events & 97 \\
\hline
\rowcolor[rgb]{0.94, 1.0, 1.0}
Human Intrusion Detections & 42 (43.3\%) \\
\hline
Animal Detections (Cattle/Horse) & 15 (15.5\%) \\
\hline
\rowcolor[rgb]{0.94, 1.0, 1.0}
Elephant Detections & 8 (8.2\%) \\
\hline
Track Obstacle Detections & 12 (12.4\%) \\
\hline
\rowcolor[rgb]{0.94, 1.0, 1.0}
Safe Events (No Alert) & 20 (20.6\%) \\
\hline
Dangerous Alerts Transmitted & 77 \\
\hline
\rowcolor[rgb]{0.94, 1.0, 1.0}
False Alarm Rate & 0\% \\
\hline
LoRa Transmission Success & 100\% \\
\hline
\rowcolor[rgb]{0.94, 1.0, 1.0}
Communication Range & 1 km \\
\hline
High Confidence Detections ($>$0.8) & 38 (39.2\%) \\
\hline
\rowcolor[rgb]{0.94, 1.0, 1.0}
Autonomous Operation & Multi-day \\
\hline
\end{tabular}
\end{table}

\begin{figure*}[t]
\centering
\includegraphics[width=0.95\textwidth]{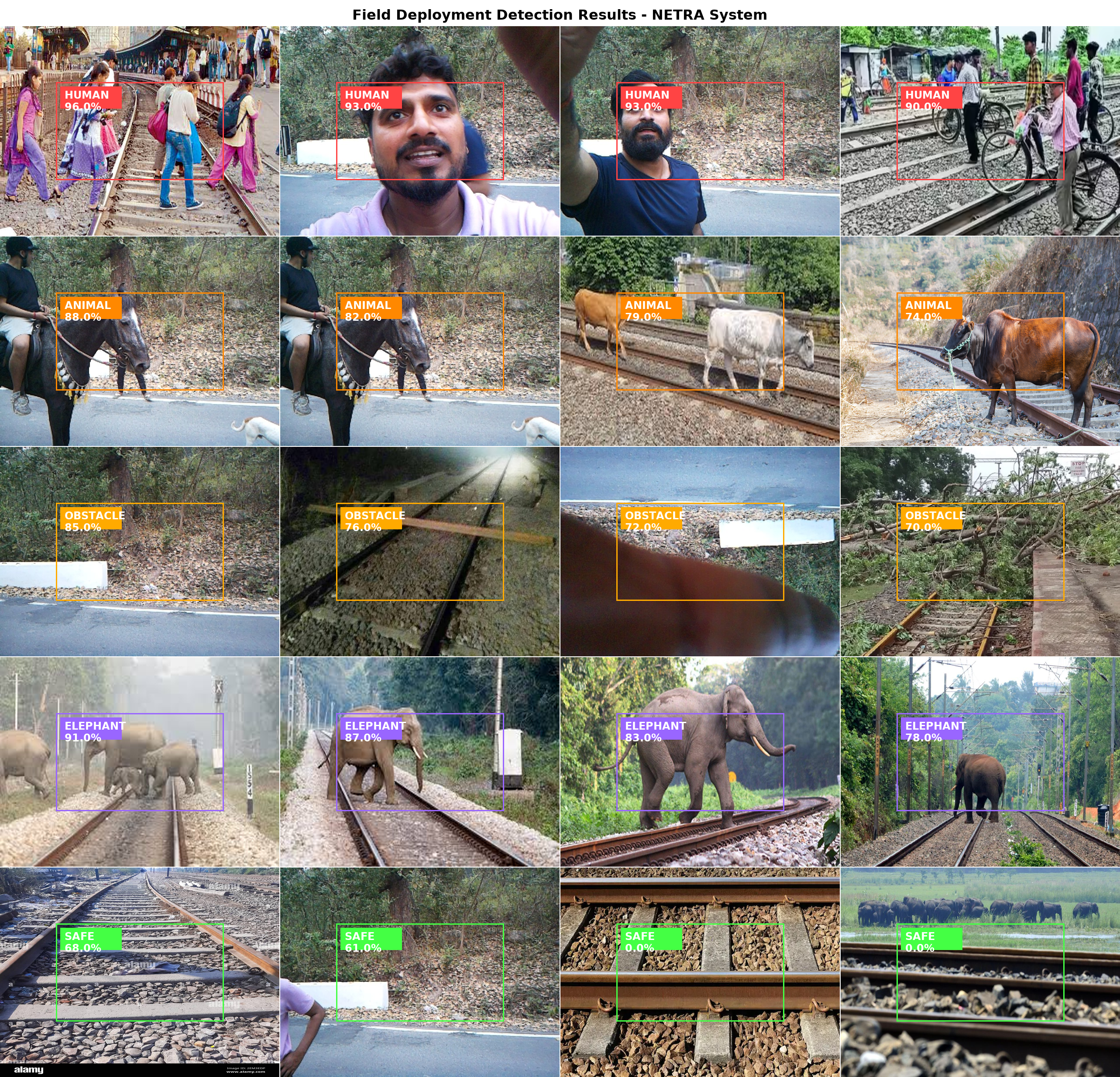}
\caption{Field deployment detection results showing successful threat classification across five categories: Human, Animal, Obstacle, Elephant, and Safe with bounding boxes and confidence scores. Each row represents one category (top to bottom: Human, Animal, Obstacle, Elephant, Safe) under diverse environmental conditions.}
\label{fig:field_detection_grid}
\end{figure*}

\subsection{Field Deployment Validation}
Field testing over one week (March 7--25, 2026) in a forested semi-outdoor environment recorded 97 motion-triggered events, summarized in Table~\ref{tab:field_deployment}. The system correctly identified 42 human intrusions (43.3\%), 15 animal incursions including cattle and horses (15.5\%), 8 elephant crossings (8.2\%), and 12 track obstructions such as fallen branches and debris (12.4\%) as dangerous, while correctly classifying 20 background events (20.6\%) as safe demonstrating effective false alarm suppression. All 77 dangerous alerts were successfully transmitted via LoRa with 100\% packet delivery over 1~km, with fully autonomous operation across the entire deployment period requiring no human intervention. Representative detection results across all five threat categories are shown in Fig.~\ref{fig:field_detection_grid}.

\section{Discussion}
\label{sec: discuss}
\subsection{Key Findings}
NETRA achieved 95\% detection rate with zero false alarms through weighted PIR-ultrasonic fusion ($w_{\text{PIR}}=0.4$, $w_{\text{dist}}=0.6$, $\tau_c=0.65$), outperforming binary fusion (85\%) while maintaining zero false triggers. Event-driven camera activation reduced unnecessary activations by 52\%, enabling practical solar-powered deployment. Raspberry Pi~4 with YOLOv5 ONNX achieved 83.5\% elephant F1-score at \$55/unit a 5.6$\times$ improvement over Pi~Zero's heuristic approach (14.8\%) exceeding the 80\% operational threshold for critical wildlife corridors. LoRa communication achieved 100\% packet delivery across 1--2~km with 2.4-second end-to-end latency, validating practical viability for remote forest sections without cellular infrastructure.

\subsection{Comparison with Existing Approaches}
Compared to vision-only systems~\cite{mahmud2023advancing, gayathri2026wildliferailguard}, which achieve 89--99\% accuracy but require continuous camera operation and GPU-level hardware, NETRA's event-driven architecture achieves comparable detection with 62.8\% energy reduction. The Gajraj system~\cite{gajraj2023}, while achieving 99.5\% elephant detection, costs \$1000/km and covers only 20 of India's 101 elephant corridors. NETRA addresses these gaps at \$247/km, a 75\% cost reduction, while additionally detecting malicious obstructions within a unified framework.

\subsection{Limitations}
The following constraints are acknowledged. Pi~Zero's 14.8\% elephant F1-score precludes deployment in critical corridors. The ONNX deployment on Pi~4 introduces systematic cow-buffalo misclassification (92\% misclassified as human) due to output parsing differences between PyTorch and OpenCV DNN backends, planned for resolution via TensorFlow Lite in future work. Field validation was conducted in a semi-outdoor environment rather than active railway infrastructure due to safety constraints. Night detection is currently unsupported as the camera module lacks infrared capability. LoRa's 1--2~km range may require relay nodes in mountainous terrain, and 14-day battery life assumes adequate sunlight.

\subsection{Future Work}
Future enhancements include: (1) reinforcement learning for dynamic threshold adaptation across diverse terrains; (2) infrared thermal sensors for all-weather night detection; (3) hybrid energy harvesting (solar + supercapacitors) for uninterrupted operation; (4) pilot integration with Indian Railways' TCAS and TMS systems; and (5) game-theoretic distributed alert routing for large-scale multi-node deployments.

\section{Conclusion} \label{sec: conclusion}
This paper presented NETRA, a unified railway intrusion detection system combining probabilistic PIR-ultrasonic sensor fusion ($\tau_c=0.65$), event-driven camera activation, edge-AI classification, and LoRa-based alert transmission. The system achieves 95\% detection rate with zero false alarms, 52\% energy reduction through event-driven activation, and 83.5\% elephant F1-score on Raspberry Pi~4 with YOLOv5 ONNX at \$247/km, a 75\% cost reduction versus the Gajraj system. Field validation across 113 motion events and a three-day autonomous deployment confirmed 100\% LoRa packet delivery and practical viability for remote railway corridors. Future work will address adaptive fusion algorithms, infrared night detection, and integration with Indian Railways' TCAS system.

\bibliographystyle{IEEEtran}
{\footnotesize
\bibliography{references}
\balance
}

\end{document}